\documentclass{ecai}
\usepackage{times}
\usepackage{graphicx}
\usepackage{latexsym}
\usepackage{lipsum}
\usepackage{multirow}
\usepackage{color}
\usepackage{amsmath}
\usepackage{fancyhdr}
\ecaisubmission   % inserts page numbers. Use only for submission of paper.
\newcommand{\ie}{\textit{i}.\textit{e}., }
\newcommand{\eg}{\textit{e}.\textit{g}. }
\newcommand{\myparagraph}[1]{\textbf{#1} -- }
\newcommand{\mysubparagraph}[1]{\textit{#1} - } 

\newcommand\mydots{\makebox[0.7em][c]{.\hfil.\hfil.}}

\begin{document}

\title{Weak Supervision helps Emergence of Word-Object Alignment and improves Vision-Language Tasks}

\author{Corentin Kervadec \institute{Universit\'e de Lyon, INSA-Lyon, LIRIS UMR
CNRS 5205, Villeurbanne, France; Orange Labs, Cesson-S\'evign\'e, France; corentin.kervadec@orange.com} \and 
Grigory Antipov \institute{Orange Labs, Cesson-S\'evigne, France; grigory.antipov@orange.com} \and 
Moez Baccouche \institute{Orange Labs, Cesson-S\'evigne, France; moez.baccouche@orange.com} \and 
Christian Wolf \institute{Universit\'e de Lyon, INSA-Lyon, LIRIS UMR CNRS 5205,
Villeurbanne, France; christian.wolf@insa-lyon.fr} }

\maketitle
\bibliographystyle{ecai}

\begin{abstract}
The large adoption of the self-attention (i.e. transformer model) and
BERT-like training principles has recently resulted in a number of high
performing models on a large panoply of vision-and-language problems (such as
Visual Question Answering (VQA), image retrieval, etc.). In this paper we claim
that these State-Of-The-Art (SOTA) approaches perform reasonably well in structuring
information inside a single modality but, despite their impressive
performances, they tend to struggle to identify fine-grained inter-modality
relationships. Indeed, such relations are frequently assumed to be implicitly
learned during training from application-specific losses, mostly
cross-entropy for classification. While most recent works provide inductive
bias for inter-modality relationships via cross attention modules, in this
work, we demonstrate (1) that the latter assumption does not hold, i.e.
modality alignment does not necessarily emerge automatically, and (2) that
adding weak supervision for alignment between visual objects and words
improves the quality of the learned models on tasks requiring reasoning. In
particular, we integrate an object-word alignment loss into SOTA
vision-language reasoning models and evaluate it on two tasks – VQA and
Language-driven Comparison of Images. We show
that the proposed fine-grained inter-modality supervision significantly
improves performance on both tasks. In particular, this new learning signal allows
obtaining SOTA-level performances on GQA dataset (VQA task) with pre-trained models without
finetuning on the task, and a new SOTA on NLVR2 dataset (Language-driven Comparison of Images). Finally, we also illustrate the impact of the
contribution on the model’s reasoning by visualizing attention distributions.
\end{abstract}

\section{Introduction}
\begin{figure}[t]
   \begin{center}
      \includegraphics[width=1.0\linewidth]{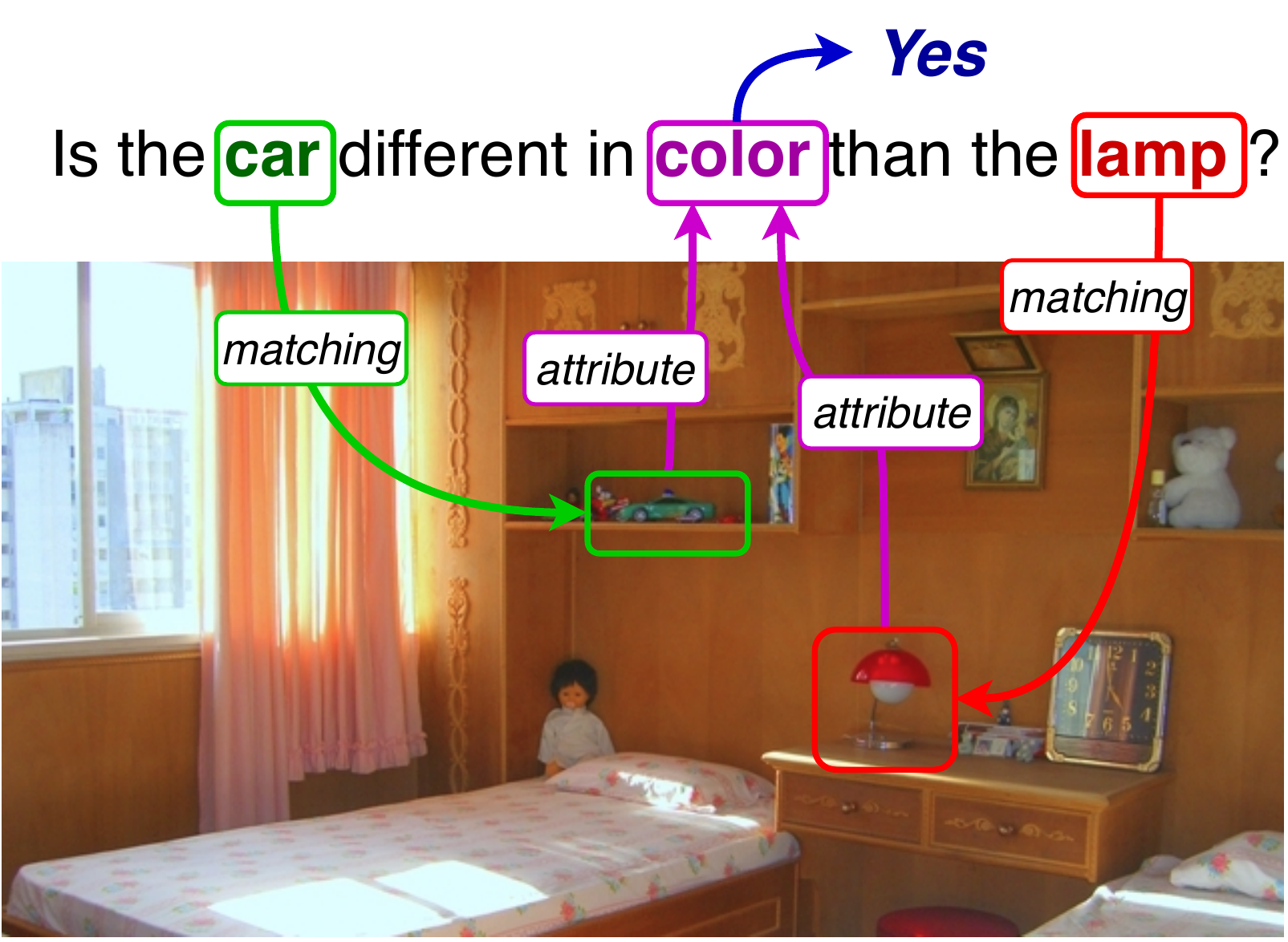}
   \end{center}
      \caption{In the context of vision+language problems, we show that the
      alignment of visual objects to words in an input sentence does not
      naturally emerge from task oriented losses and propose a new auxiliary
      training objective addressing this problem. The presented image and question are taken
      from GQA~\cite{Hudson_2019_CVPR} dataset.}
   \label{fig:alignement_supervision}
\end{figure}

High-capacity deep neural networks trained on large amount of data currently
dominate methods addressing problems involving either vision \textit{or}
language, or both of these modalities jointly. Examples for vision-language
tasks are image retrieval task~\cite{karpathy2015deep} (retrieve an image given a query sentence); image
captioning~\cite{lin2014microsoft} (describe the content of an input image in one or more sentences),
and Visual Question Answering~\cite{antol2015vqa} (VQA: textually answer a question on an input
image) \textit{etc}.
These tasks require different forms of reasoning, among which we find the capacity to analyze instructions -- \eg 
the question in VQA --, or the ability to fuse modalities or to translate one modality into another one -- \eg in image captioning.
Additionally, they often require different levels of understanding, from a global image-text comparison to
 fine-grained object-word matchings.

In this context, a large panoply of high-performing models adopt
self-attention architectures \cite{vaswani2017attention} and BERT-like
\cite{devlin2019bert} training objectives, which complement the main task-related loss with other auxiliairy losses correlated to the task.
% recently emerged for vision and language problems. 
The common point of this large body of work is the large-scale training of unified vision-language encoders on image-sentence pairs.
However, despite their ability to model interactions unique to one modality (i.e. \textit{intra}-relationships),
we observe that these State-Of-The-Art (SOTA) approaches tend to struggle to identify fine-grained object-word relationships (\textit{inter}-relationships, or cross-modality relationships). 
These relationships are important, which can be illustrated in the example of
VQA: answering a question given an input image requires the detection of certain
objects in the image, which correspond to words in the question, and eventually
the detection of more fine-grained relationships between visual objects, which
are related to entities in the sentence.

In the literature, the alignment or matching of words to visual objects is generally assumed to be implicitly learned from application-specific losses ---
mostly cross-entropy for classification --- thanks to the inductive biases provided by the encoder's architecture, i.e. the possibility of the model to \emph{represent} this kind of matching.
In this work we show that (1) modality alignment (\textit{cf}. Figure~\ref{fig:alignement_supervision}) does not necessarily emerge automatically and
(2) that adding weak supervision for alignment between visual objects and words
improves the quality of the learned models on tasks requiring visual reasoning.

Our contributions are as follows:
\begin{itemize}
   \item  We enhance vision-language encoder approaches by adding explicit weak
   supervision of object-word alignment, taking into account the uncertainty 
   present in the detection result of the vision module.
   
   \item We improve the accuracy of SOTA vision-language models on the VQA task
   GQA \cite{Hudson_2019_CVPR} dataset) \emph{without the need of finetuning}, to achieve
   SOTA-level results. In other words, with our new objective, pre-training is
   sufficient for SOTA results.
   
   \item On the task of Language-driven Comparison of Images, requiring to reason over two images and one sentence, our proposed model outperforms the current SOTA model on the challenging NLVR2 dataset.
   
   \item We show visualizations of attention maps, which corroborate the claim that word-object alignment does not naturally emerge from task losses, while it is discovered by our weak supervision signal.
   
\end{itemize}

\section{Related Works}
\label{sec:related_works}

\myparagraph{Vision-language tasks}
% Our work contribute to improve vision-language models. By vision-language
% model, we understand models built to solve tasks requiring to understand both
% vision and language. 
\textit{Vision and language understanding} is a broad
area and can take several forms at many different levels of granularity.
Some tasks focus on matching problems, as for instance \emph{Image Retrieval}, which 
requires finding the most relevant image given a query sentence
\cite{karpathy2015deep}, \cite{lee2018stacked}. The inverse problem
--- namely \emph{Sentence Retrieval} --- has also been explored \cite{karpathy2015deep}. A similar task with finer
granularity is \emph{Visual Grounding}, where the model must associate image regions 
to words or sentences \cite{kazemzadeh2014referitgame}, \cite{plummer2015flickr30k}.

Other tasks require more high-level reasoning over images and sentences, which, in general, requires multi-modal interactions but also the ability to compare, count or find relations between objects in the image.
In \emph{Visual Question Answering} (VQA) \cite{antol2015vqa} \cite{Hudson_2019_CVPR} we ask questions (given as input) about an input image and the model
must predict the answer. Answering the questions requires a variety of skills: finding relations,
counting, comparing colors or other visual features, materials, sizes, shapes, \textit{etc}.  
The binary task of \textit{Language-driven Comparison of Images} takes as input triplets $(img_1,img_2,sentence)$
and requires predicting whether the sentence truly describes the image pair \cite{suhr2019corpus}. 
%\emph{Natural Language for Visual Reasoning for Real} (NLVR2)
% However it goes
% beyond a classic visual language comparison. Indeed the dataset is built so as to involve comparison,
% counting, relation prediction, etc.
% Faut il parler de VCR?

Finally, some tasks involve the generation of one modality from the other. \emph{Image captioning} consists in 
translating an image into text \cite{lin2014microsoft}. Similarly, some tasks aim to generate
questions about an image \cite{li2018visual}. Inversely, it is also possible to generate an image from a caption \cite{mansimov2015generating}.
However, such multimodal generation is out of the scope of our work.

\myparagraph{Vision-language multi-modal fusion}
Early work in vision and language understanding focused on separate models for
each modality followed by multi-modal fusion \cite{ren2015exploring}. In this context, bi-linear fusion
is an expressive family of models, which, however, suffers from
overparametrization and therefore overfitting. Subsequent work addressed this by
creating low-rank decompositions of the fusion tensors, either through Tucker
tensor compositions as in MUTAN~\cite{ben2017mutan}, or block tensor decompositions like in
BLOCK~\cite{ben2019block}.
% \myparagraph{Towards holistic architectures}

However the general tendency is to move towards holistic architectures,
modeling all the interactions between modalities, and also between different objects in the visual modality. 
Object level reasoning, i.e. the analysis of visual data in the form of a collection of previously detected 
local entities/objects, has become a general tendency in computer vision beyond VQA, also seen in 
video analysis \cite{BaradelECCV2018} etc.
In this context, the Relation Network \cite{santoro2017simple} considers all the pairwise interactions between
visual objects.
\cite{teney2017graph}, \cite{norcliffe2018learning} and \cite{lu2019vilbert}
apply variants of Graph Convolutional Network \cite{kipf2016semi} on visual objects and question words
for VQA. \cite{yu2019deep} and \cite{gao2019dynamic} go a step further by
modeling multimodal interactions via adapting transformer
\cite{vaswani2017attention} principles to vision and language.
We call them holistic because they consider both intra-modality (inside a modality) 
and inter-modality (fusion between modalities) relationships.

% Put these references in the previous myparagraph
% \begin{itemize}
%   \item CNN + LSTM: The two modalities are processed independantly -- e.g. using CNN
%   (resp. LSTM) for vision (resp. question) -- then fused using more or less
%   complex rules. Ex: Hadamard, tensor fusion (MUTAN, BLOCK)
%   \item Bottom Up Top Down: There is no intra relations, but the relation
%   between the whole words and each visual object is used as to weights visual
%   objects.
%   \item Relation Network: Intra relations between visual object is computed
%   within the RN module. Visual object pairs are also fused with the whole text
%   representation.
%   \item MUREL improves the RN module by fusing question and visual objects
%   using tensor fusion.
%   \item Graph networks and co.: graph network approaches tries to make use of
%   the intra modality relations between objects (others references: LCGN, Learning Conditioned Graph
%   Structures).
%   \item Finally, recent models as MCAN or DFAF consider all intra modal and
%   inter modal relations. They generalize graph networks approaches in an
%   efficient way.     
% \end{itemize}

\myparagraph{Multi-task pretraining}
A second tendency is the evolution of training from 
task-specific supervision signals to a set of different losses,
which are related to general vision-language understanding, and whose supervision signal can successfully be transferred to different downstream tasks. 

This use of auxiliary tasks and knowledge transfer can be performed on both modalities: 
on language, the use of word embeddings such as
GloVe~\cite{pennington2014glove} or BERT~\cite{devlin2019bert} is frequent. On vision, usual objectives include 
the use of pre-trained object detectors such
as Faster RCNN~\cite{ren2015faster}, BUTDé\cite{anderson2018bottom} for VQA and image captioning, and SCAN~\cite{lee2018stacked} for image retrieval.

More recent work shows that a joint pre-training over both modalities can benefit 
downstream vision-language tasks.
This is achieved by setting up strategies to learn a
vision-language representation in a multi-task fashion similar to
BERT~\cite{devlin2019bert} in Natural Language Processing (NLP).
Thereby, LXMERT~\cite{tan2019lxmert} and VilBERT~\cite{lu2019vilbert} use holistic architectures to learn a
vision-language encoder trained on a large-scale amount of images-sentences pairs. Their encoder
is then transferred to specific vision-language tasks, where they generally achieve
SOTA results.

\myparagraph{Symbolic representation for visual reasoning}
Aside from these approaches, others address the visual reasoning problem by constructing a
symbolic view of vision, language and of the reasoning process. 
Thus, \cite{yi2018neural} uses reinforcement learning to learn a program
generator predicting a functional program from a given question in order to model the reasoning.
NSM~\cite{hudson2019learning} predicts a probabilistic graph from the image to
obtain an abstract latent space which is then processed as a state machine.

%\paragraph*{}
Our work follows tendencies in SOTA and is based on a holistic joint
vision-language encoder with multiple training objectives.
We improve on the SOTA by adding an additional objective, 
which (weakly) supervises the model to align objects and words referring to the same entity.
This new objective complements the inductive bias inherent in current models, which allows learning of cross-modality relationships.

\section{Vision-Language Encoder}
\label{sec:vl_encoder}

\begin{figure*}[t!]
   \begin{center}
      \includegraphics[width=1.05\linewidth]{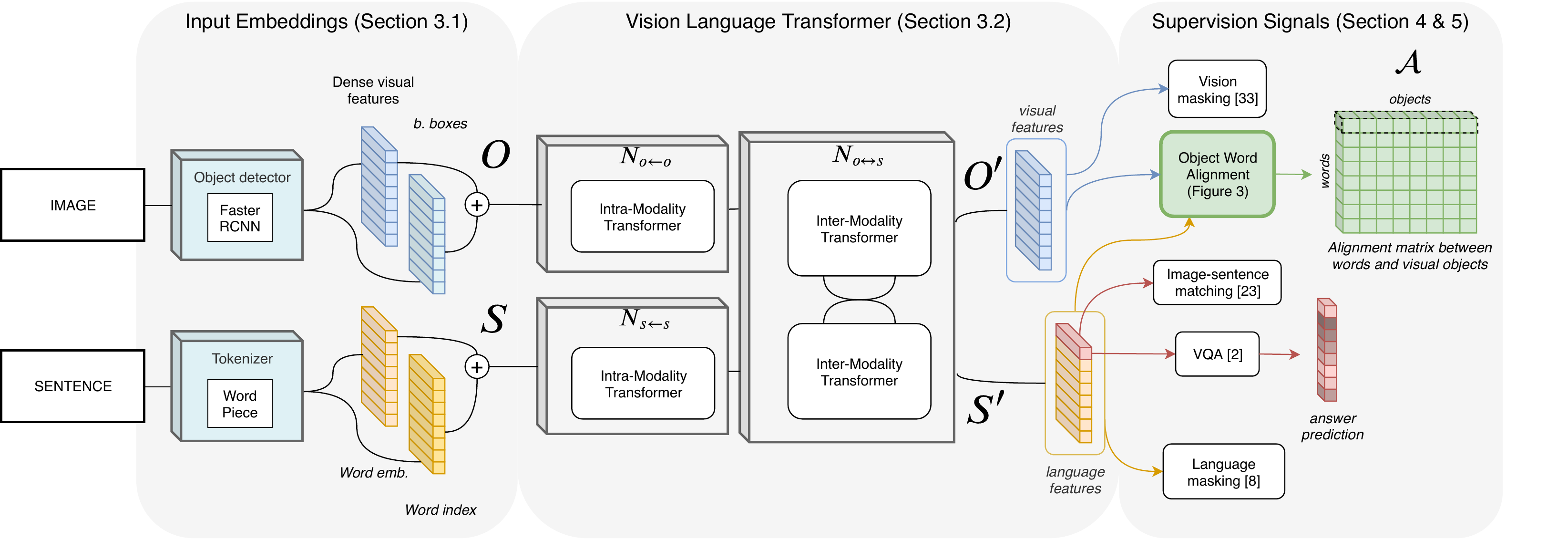}
   \end{center}
      \caption{Architecture of our vision-language encoder (Section~\ref{sec:vl_encoder}) and the respective
      supervision tasks (Sections~\ref{sec:supervision_signals} and~\ref{sec:alignment}).
      From left to right: (left), image and sentence are embedded into \textit{input
      embeddings} $(O, S)$ (Section~\ref{subsec:input});
      (middle), they are then processed by the \textit{vision-language transformer}
      (Section~\ref{subsec:vl_transformer}) modeling both intra- and
      inter-modality interactions in order to obtain multimodal embeddings $(O',
      S')$;
      (right), the encoder is supervised using objectives introduced in
      Section~\ref{sec:supervision_signals} and our soft object-word alignment
      proposed in Section~\ref{sec:alignment}.
      }
   \label{fig:overall}
\end{figure*}

\noindent
% We build on and extend state-of-the-art vision-language encoders, in particular we follow \cite{yu2019deep}.
% A block-diagram is given in Figure~\ref{fig:overall}, the details will be
% developed in what follows. The different training objectives will be explained
% in subsequent sections.
In this Section, we present our vision-language encoder which is used for
learning multimodal embeddings. The encoder is built upon the recent work, and
in particular~\cite{yu2019deep}. The overall architecure of our model is
presented in Figure~\ref{fig:overall}. Below, we firstly present the embeddings
extraction part of our encoder (Section~\ref{subsec:input}), and then focus on
its central part (Section~\ref{subsec:vl_transformer}).

\subsection{Input Embeddings}
\label{subsec:input}

\myparagraph{Vision Input}
On the vision side, we use an object detector -- Faster-RCNN~\cite{ren2015faster} -- to extract object level-visual
features from the input image as in \cite{anderson2018bottom}.
Similar to hard attention mechanisms, this enforces the system
to reason on object level rather than on the pixel level or global level.
Thus, for each image we extract $N_o{=}36$ bounding boxes and associated 2048-dimensional visual features:
\begin{equation}
   % [[f_0, \dots, f_{N_o-1}], [b_0, \dots, b_{N_o-1}]] = \textit{RCNN}(I),
   (f,b) = \textit{RCNN}(I),
\end{equation}
where $I$ is the input image and $f{=}\{f_0, \mydots, f_{N_o-1}\}$ and $b{=}\{b_0, \mydots, b_{N_o-1}\}$ are, respectively, the dense visual
features and the bounding boxes detected for objects. Box and dense vectors
are fused to obtain position-aware object level embeddings $O = [o_0, \dots, o_i, \dots, o_{N_o-1}]$.

\myparagraph{Language Input}
On the language side, sentences are tokenized using the WordPiece tokenizer \cite{wu2016google}.
As common in language processing, a special token \verb![CLS]! is added at the beginning of the tokenized sentence,
which encodes the multimodal information of the image and sentence. The transformation of this token, performed during the forward pass through the network, corresponds to the prediction of the answer to the task.
Tokens are embedded into $d$-dimensional vectors using a look-up table learned during a pre-training phase which concentrates on language only.
The index position of the word is added to the dense vector as a positional encoding in order to obtain 
index-aware word level embeddings $S = [s_0, \dots, s_i, \dots, s_{N_s-1}]$.

\subsection{Vision-Language Transformer}
\label{subsec:vl_transformer}

The neural model encodes the independent vision
and language embeddings $(O, S)$ described above and transforms them as they
pass through the network:

\begin{equation}
   O', S' = Encoder(O, S),
\end{equation}
where $(O', S')$ are the updated output embeddings.
We resort to the widely-used transformer architecture \cite{vaswani2017attention} adapted to
vision-language problems as in \cite{gao2019dynamic} and \cite{yu2019deep}.

The vision-language transformer is composed of two self-attention modules of type \textit{intra-modality
transformer} and \textit{inter-modality transformer} as defined in \cite{gao2019dynamic}.
They take as input one input sequence (in case of intra-modality) or two input sequences (in case of inter-modality) and calculate an output sequence:
%Beforehand, let's define the attention mechanism used in the \textit{intra} and \textit{inter} modules:
\begin{equation}
  x_i = T(q_i, k_j, v_j) = \sum_j \alpha_{ij} v_j,
\end{equation}
where $q_i$, $k_j$ and $v_j$ are, respectively the \emph{query}, \emph{key} and \emph{value}  vectors in
$R^d$ \cite{vaswani2017attention}, which are calculated as linear mappings from
the input sequences. Their exact definition depends on the type (inter vs.
intra) and is given further below.

The $\alpha_{ij}$ represent an attention map, which predicts how the different
elements of the input sequences attend to each other:
\begin{equation}
   \label{eq:attention_maps}
   \alpha_{ij} = softmax_i( \frac{q_i^Tk_j}{\sqrt{d}} ),
\end{equation}
where $d$ is the number of dimensions of the embedding space. As in \cite{vaswani2017attention}, we use multi-head attention, where the embeddings are split into $H$ parts, transformations are calculated in parallel, and predictions concatenated. Each transformer layer is followed by a residual
connection, a layer normalization~\cite{ba2016layer}, and a feed-forward layer.

\myparagraph{Intra-Modality Transformer blocks} allow to model the interactions inside one
modality. Thus, their \emph{query}, \emph{key} and \emph{value}
vectors come from the same modality. They are defined as follows:
\begin{equation}
   s' = T_s(s^q, s^k, s^v)
\end{equation}
\begin{equation}
   o' = T_o(o^q, o^k, o^v),
\end{equation}
where $x^q = W^qx,$ $x^k = W^kx$ and $x^v = W^vx$. 

\myparagraph{Inter-Modality Transformer blocks} model
% \begin{figure}[t]
%    \begin{center}
%       \includegraphics[width=.8\linewidth]{cross_modal.png}
%    \end{center}
%       \caption{Crossmodal transformer. On the bottom we symbolically represents the attention maps.}
%    \label{fig:matching}
%    \end{figure}
information flowing between both modalities vision and language.
They are defined similarly to the intra-modality transformer but the \emph{key} and \emph{value}
vectors are crossed between the modalities:
\begin{eqnarray}
   s' & = & T_{s \leftarrow o}(s^q, o^k, o^v)\\
   s'' & = & T_s(s'^q, s'^k, s'^v)
\end{eqnarray}
\begin{eqnarray}
   o' & = & T_{o \leftarrow s}(o^q, s^k, s^v)\\
   o'' & = & T_o(o'^q, o'^k, o'^v)
\end{eqnarray}

\myparagraph{Stacked Architecture}
The vision-language transformer is built by stacking the previously defined modules as
shown in Figure~\ref{fig:overall}: the image/sentence input data is passed
through detectors/tokenizers, embedding extractors, several intra-modality
transformer blocks and finally several cross-modality transformer blocks. 
Summarizing, the model contains $N_{o \leftarrow o}$ and $N_{s \leftarrow s}$
stacked intra-modality transformer followed by $N_{o \leftrightarrow s}$
inter-modality transformers.

\paragraph*{}
The vision-language transformer provides inductive biases for modeling intra-modality relationships
(\eg sentence dependency graph for language, scene graphs for vision) and
inter-modality relationships (\eg vision-language alignment). However, as shown below,
inductive biases are not sufficient for learning inter-modal interactions, it is
therefore necessary to define adequate supervision signals.

\section{Supervised Objectives}
\label{sec:supervision_signals}
We train the vision-language encoder defined in Section 3
following the recently widely-adapted strategy of combining BERT-like \cite{devlin2019bert}
self-supervised signals with task-specific supervision signals, which 
has been applied to various problems
in vision and language --- \eg \cite{tan2019lxmert} \cite{lu2019vilbert}. 
% This body of work adapts the ideas of BERT \cite{devlin2019bert}, initially proposed for language tasks, to the multimodality of vision and language.
We select four supervision signals: vision masking
\cite{tan2019lxmert},
language masking \cite{devlin2019bert}, image-sentence matching
\cite{lu2019vilbert} and visual
question answering \cite{antol2015vqa}, which are briefly described below.

\myparagraph{Vision/Language Masking}
This signal aims to supervise the encoder's ability to reconstruct
missing information in language and vision. More precisely, we randomly mask 
each language token (resp. visual object) with a probability of $0.15$
and ask the model to predict the
missing words (resp. objects). Therefore we add two classifiers -- for
\textit{vision masking}\footnote{It is worth noticing that the vision masking
requires to predict both the object classes and its attributes (\eg color,
materials, \textit{etc}.)} and
\textit{language masking} -- on top of the vision language encoder and supervise
via a cross-entropy loss. \cite{tan2019lxmert} proposes to take the object
detector prediction as ground truth in order to get over the disparity of visual
annotation. Additionally, we also supervise the model to regress the
masked objects' Faster-RCNN features via L2 loss.

\myparagraph{Image-Sentence Matching}
BERT \cite{devlin2019bert} proposes \textit{next sentence prediction} supervision by
asking to predict if two sentences are consecutive in a given text, or randomly sampled from a corpus. 
Its vision-language equivalent is \textit{image-sentence matching} \cite{tan2019lxmert}
\cite{lu2019vilbert}, where the model has to predict whether
a given sentence matches a given image or not.
Thus, we randomly replace the image in each sentence-image
pair with a probability of $0.5$. We add a feed-forward layer on top of the
\verb![CLS]! output embedding to predict whether the pair matches or not. 
This global matching is
supervised using a binary cross-entropy loss. 

\myparagraph{Visual Question Answering}
Our model is applicable to a wide range of vision-language problems (in
Section~\ref{sec:experiments} we evaluate it to two different tasks, namely VQA
and Language-driven Comparison of Images). At the same time, independently of
the target vision-language task, pretraining on VQA helps reasoning as shown
in~\cite{tan2019lxmert}. The VQA task is defined as a classification problem over a set of
most frequent answers. In this work, we perform this classification from a
prediction head attached to the \verb![CLS]! token and supervise it using a
cross-entropy loss.

\section{Weak Supervision of Object-Word Alignment}
\label{sec:alignment}

The presented SOTA vision-language supervision signals -- \ie \textit{vision/language masking}, \textit{image-sentence matching} and \textit{VQA} --
have proved their efficiency at encoding rich vision-language embeddings \cite{tan2019lxmert}
\cite{lu2019vilbert}. However, none of them explicitly supervises the object-word alignment.
At the same time, matching words and visual objects referring to a same high-level entity is a natural prerequisite for visual reasoning.

The reason why such supervision has not been proposed before is probably that this fine-grained matching property is frequently assumed to be implicitly learned
via inter-modality attention modules training.
In this work, we claim that the word-object alignment does not necessarily emerge automatically
but rather requires explicit supervision.

\myparagraph{Vision-Language Alignment Decoder}
\begin{figure}[t]
   \begin{center}
      \includegraphics[width=1\linewidth]{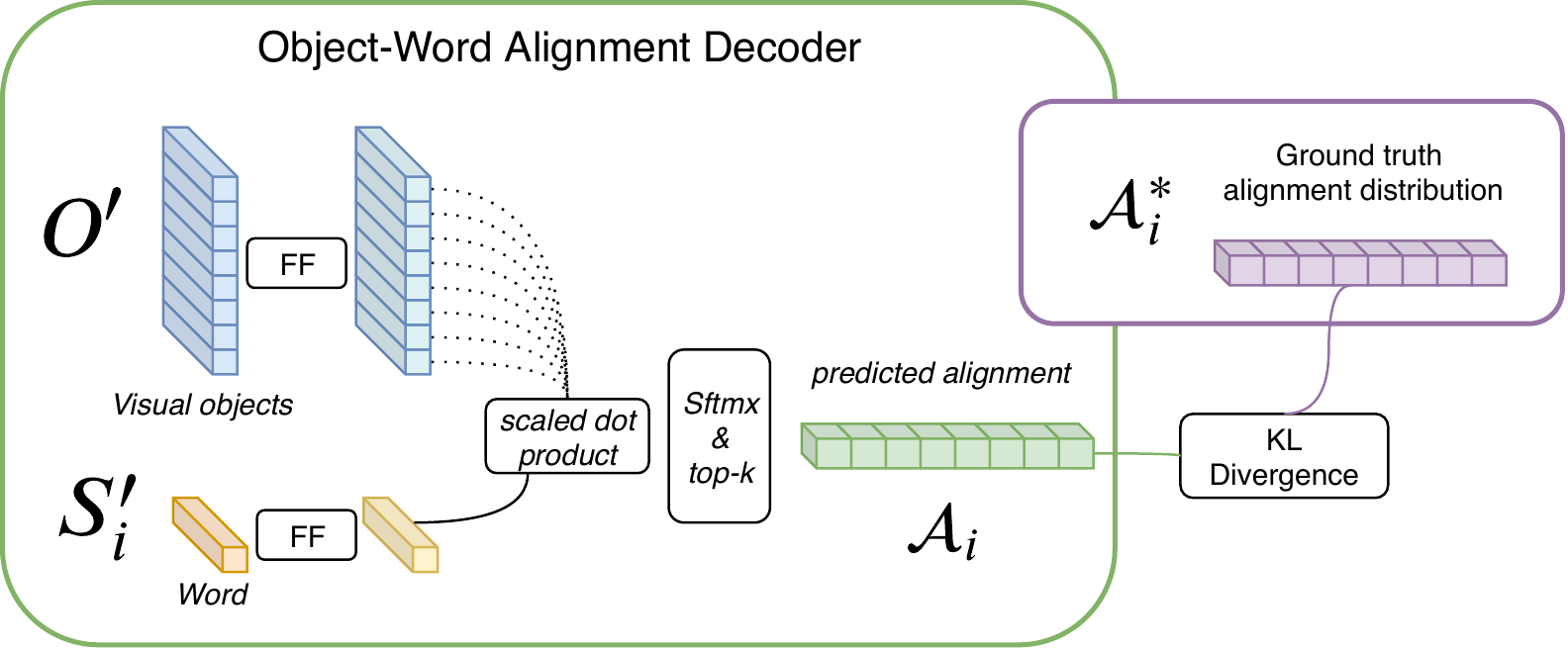}
   \end{center}
      \caption{The proposed vision-language alignment decoder and the respective
      weakly-supervised loss. In this illustration, we present the
      alignment prediction $\mathcal{A}_i$ between one
      word $S'_i$ and the visual objects $O'$.
      $FF$ stands for feed-forward layers.}
   \label{fig:matching}
\end{figure}
We propose to add a vision-language alignment decoder on top of the encoder.
The whole model is then supervised to predict the
object-word alignment matrix $\mathcal{A}$ from the encoder's outputs $(O', S')$.
First, $(O', S')$ are projected into a joint space using a feed-forward layer with layer normalization~\cite{ba2016layer} 
and residual connection. We obtain $(\hat{O}, \hat{S})$ from which we compute $\mathcal{A}$:
\begin{equation}
   \mathcal{A} = \frac{\hat{S} \otimes \hat{O}}{\sqrt{d}},
\end{equation}
where $\otimes$ is the outer product.
In other words, the alignment scalar $\mathcal{A}_{ij}$ is computed as the 
scaled-dot-product between each object-word pair $(o_i, s_j)$, as shown in Figure~\ref{fig:matching}:
\begin{equation}
   \mathcal{A}_{ij} = \frac{\hat{s}_i . \hat{o}_j^T}{\sqrt{d}}
\end{equation}
For each word $s_i$ we only keep the top-$k$ highest predictions and apply a softmax:
\begin{equation}
   \mathcal{A}_i = softmax_j(top_k(\mathcal{A}_{ij}))
\end{equation}
In this work, we empirically set $k=3$.
This way, we compute from each word a probability distribution $\mathcal{A}_i$ over the set of
visual objects detected by Faster-RCNN. A high probability $\mathcal{A}_{ij}$ means word $s_i$ and object $o_j$ refer to
the same high-level entity. 
The dedicated loss $L_{align}$ is defined using Kullback-Leibler ($KL$) divergence:
\begin{equation}
   L_{align} = KL(\mathcal{A}^*, \mathcal{A}),
   \label{eq:align_loss}
\end{equation}
%\begin{equation}
%   L_{pointer} = \frac{1}{n_{words}}\sum_i^{n_{words}} m^*_i \cdot (log \ m^*_i - log \ m_i)
%   \label{eq:pointer_loss}In order to feed our encoder with object-word alignment supervision, we gather annotations from GQA and Visual Genome.
\myparagraph{Soft Alignment Score: approximating $\mathcal{A}^*$}%or Dealing with the detection uncertainty}
Let's suppose we have the ground truth object-word pair $(s_i, b^*_{s_i})$ (\textit{cf}. Section~\ref{sec:data}).
This pair is composed of a word or group of words $s_i$ taken from the input sentence and a bounding box
$b^*_{s_i}$ indicating the position of the respective object in the image.
However we cannot directly use this supervision because both ground truth
object-word annotations and the object detector are imperfect. More precisely, (1) the ground
truth visual-object annotation is often misaligned with the object detection's
bounding box prediction, or
(2) the annotated object can simply be not detected at all. 
% For instance, the object detector detects the correct object but not at the very same position as in the annotation -- case (1);
% the object detector does not detect an object provided in the annotation -- case (2);
% the object detector mislabels a correctly detected object -- case (2); \textit{etc}.   

To address this issue we set up a soft-alignment score taking into account both the detection-annotation misalignment 
and the object detector imperfection. To this end, we consider two criteria: the position one and the semantic one.

\mysubparagraph{Position Criterion}
For each ground truth object-word pair $(s_i, b^*_{s_i})$, we compute Intersection over Union (IoU)
between object detector's predicted bounding box $b_{o_j}$ and the ground truth object's bounding box $b^*_{s_i}$:
\begin{equation} 
   P\mathcal{A}^*_{ij} = IoU(b^*_{s_i}, b_{o_j}),
\end{equation}
A high IoU leads to a high criterion value. Therefore, this criterion permits to
give more importance to objects detected in the same image region as the
ground-truth object.

\mysubparagraph{Semantic Criterion}
At the same time, we cannot only rely on positional information. Indeed, we also have
to take into account the semantics of the object detector's prediction. This would
avoid to align a word with a well-localized but a semantically-different object (according to the detector).
Therefore we define the semantic criterion which computes the semantic similarity between a word $s_i$ and the object's
class $c_{o_j}$ -- and attribute $a_{o_j}$ -- predicted by the detector:
\begin{equation} 
   S\mathcal{A}^*_{ij} = \frac{3}{4}S(s_i, c_{o_j}) + \frac{1}{4}S(s_i, a_{o_j}),
\end{equation}
where $S(\cdot,\cdot)$ compute the cosine similarity between the GloVe embeddings of the
class/attribute names. We bias the similarity toward object class as we empirically found it more relevant than the
attribute prediction.

\paragraph*{}
Finally, we combine the two criteria in order to obtain a soft alignment score for each object-word pair in the annotation:
\begin{equation}
   \mathcal{A}^*_{ij} = \frac{norm_j(P\mathcal{A}^*_{ij}) + norm_j(S\mathcal{A}^*_{ij})}{2}
\end{equation}
The resulting soft-alignment scores are normalized over the objects such as:
\begin{equation}
   \sum_j^{n_{objects}} \mathcal{A}^*_{ij} = 1
\end{equation}
Hence the ground truth soft alignment score $\mathcal{A}^*_i$ of word $s_i$ is a probability distribution over the set of
visual objects detected by the object detector.

The soft alignment
score defined in this Section is by construction approximate. Therefore, we refer to the designed
supervision signal as weak.

\section{Experiments}
\label{sec:experiments}
In this section we evaluate the learned vision-language encoder on two tasks, namely
VQA \cite{antol2015vqa} and the Language-driven Comparison of Images~\cite{suhr2019corpus}.
%We first start by describing our training data with a focus on the alignment annotation.

\begin{table*}
   \begin{center}
   {\caption{Evaluation of the proposed object-word alignment weak supervision
   on the GQA~\cite{Hudson_2019_CVPR} dataset. The presented results are
   calculated on the dataset's test-std split. The GQA's accuracy is presented
   in the last column. The exact definitions of all other (auxiliary) metrics
   can be found in~\cite{Hudson_2019_CVPR}.}\label{table:VQA_SOTA}}
   \begin{tabular}{lccccccc}
   \hline
   \rule{0pt}{12pt}
   \rule{0pt}{12pt}
   Models&Binary&Open&Validity&Plausibility&Consistency&Distribution&Acc.\\
   \hline
   \\[-6pt]
   Human \cite{Hudson_2019_CVPR} & 91.2 & 87.4 & 98.9 &  97.2 & 98.4 & - & 89.3\\
   BUTD \cite{anderson2018bottom} & 66.6 & 34.8 & 96.2 & 84.6 & 78.7 & 6.0 & 49.7\\
   MAC \cite{hudson2018compositional} & 71.23 & 38.9 & 96.2 & 84.5 & 81.6 & 5.3 & 54.1\\
   LCGN \cite{hu2019language} & 73.7 & 42.3 & \textbf{96.5} & \textbf{84.8} & 84.7 & 4.7 & 57.0\\
   LXMERT \cite{tan2019lxmert} & 77.2 & 45.5 & 96.4 & 84.5 & 89.6 & 5.7 & 60.3\\
   NSM \cite{hudson2019learning} & \textbf{78.9} & \textbf{49.3} & 96.4 & 84.3 & \textbf{93.3} & \textbf{3.7} & \textbf{63.2}\\
   \hline
   \rule{0pt}{8pt}
   \textbf{ours} & \textbf{76.9} & \textbf{46.1} & \textbf{96.3} & \textbf{84.7} & \textbf{89.7} & \textbf{5.3} & \textbf{60.5}
   \\
   \hline
   \\[-6pt]
   \end{tabular}
   \end{center}
\end{table*}

\subsection{Training Data}
\label{sec:data}

% \begin{table*}
%    \begin{center}
%    {\caption{Our training data (based on \cite{tan2019lxmert} Table~1).
%    The last row indicates the number of sentences annotated with object-word pointers, along with the
%    proportion regarding to the total number of sentences.}\label{table:data}.
%    }
%    \begin{tabular}{|l|c|cccccc||c|}
%    \hline
%    \rule{0pt}{12pt}
%    Database&Images&COCO-Captions&VG-Captions&VQA 2.0&GQA&VG-QA&All&Object-Word Pointer\\
%    \hline
%    \\[-6pt]
%    MSCOCO - VGENOME& 72K & 361K & - & 387K & - & - & 0.75M & 0 (0\%)\\
%    MSCOCO $\cap$ VGENOME& 51K & 256K & 2.54M & 271K & 515K & 724K & 4.30M & 3.06M (33\%)\\
%    VGENOME - MSCOCO& 57K & - & 2.85M & - & 556K & 718K & 4.13M & 3.40M (37\%)\\
%    All& 180K & 617K & 5.39M & 658K & 1.07M & 1.44M & 9.18M & 6.46M (70\%)
%    \\
%    \hline
%    \end{tabular}
%    \end{center}
% \end{table*}

Following \cite{tan2019lxmert}, we construct our dataset as the concatenation of
the two public ones, namely: MSCOCO~\cite{lin2014microsoft} and Visual Genome~\cite{krishna2017visual}. These datasets provide images annotated with captions.
The VQA annotations are taken from three datasets (based on images from either MSCOCO or Visual Genome):
VQA~2.0~\cite{balanced_vqa_v2}, GQA~\cite{Hudson_2019_CVPR} and VG-QA~\cite{krishna2017visual}.
Consequently, our dataset is composed of 9.18M image-sentence pairs
(a sentence can be either a caption or a question).

The object-word alignment scores, which are defined in
Section~\ref{sec:alignment}, are calculated based on the annotations extracted
from GQA and Visual Genome.
In GQA dataset, salient question words and answers are annotated with visual
pointers. A visual pointer consists in a bounding box corresponding to the visual
region described by the words composing the question or the answer.
Nevertheless, as GQA represents only 12\% of the dataset, the use of the GQA
pointers would have been insufficient.

To alleviate this issue, we augment the pointer annotation with visual
grounded annotations from Visual Genome.
Every Visual Genome image is accompanied with visual region
descriptions forming \textit{(description, bounding box)} pairs. Unlike in GQA,
descriptions are full descriptive sentences and not small groups of words. Therefore, the so
obtained pointer is less discriminative towards the language part. 
Therefore, we choose to combine these descriptions in order to obtain sentences with
one, two or three pointers. For instance the two descriptions \textit{"the cat playing
near the tree"} and \textit{"the yellow bird"} become \textit{"the cat playing near the tree
and the yellow bird"}, with the associated bounding boxes. 

All in all, by combining annotations from GQA and Visual Genome, we gather roughly $6$M image-sentence pairs annotated with pointers.
In other words, about $70$\% of the total number of the image-sentence pairs in the dataset have fine-grained object-word alignment annotations.
% Table~\ref{table:data} provides all the details
% about how our dataset is constructed.

\subsection{Evaluation Tasks}

To evaluate the reasoning quality of our vision-language encoder supervised with
the object-word alignment, we evaluate it on two tasks requiring to reason over image and text.

The first one is the VQA task. 
It consists in predicting the answer asked about an image.
This task is challenging as it requires a high-level understanding of 
vision and language.
For evaluation, we select the most recent and, arguably, the most challenging VQA-dataset today, namely GQA.
As GQA is already used during the vision-language encoder training,
we do not find it necessary to finetune our model on the datatset. 

The second is the Language-driven Comparison of Images.
We choose the Natural Language for Visual Reasoning (NLVR2) dataset \cite{suhr2019corpus}. NLVR2 is composed of triplets
$(img_1,img_2,sentence)$ where $img_1$ and $img_2$ are two images and $sentence$ is a
sentence describing one or both images. The goal is to predict if the sentence is
true. 
It is worth noticing that NLVR2 data is not viewed during the encoder training, therefore it truly
evaluates the generalization capacity of our method.
We use the same finetuning strategy as in \cite{tan2019lxmert}. Thus we concatenate the two encoder's output $[CLS]$
embeddings -- obtained with $(img_1,sentence)$ and $(img_2,sentence)$ pairs -- and pass
them through a feed-forward layer. We then use a binary cross-entropy loss.

\subsection{Results}

\myparagraph{Training Details}
We train our vision language encoder using Adam optimizer
\cite{kingma2014adam} during 20 epochs. However, the VQA supervision is only added
after 10 epochs, following \cite{tan2019lxmert}. 
We set the learning to $10^{-4}$ with warm starting and learning rate decay.
The batch size is 512.
On the architecure side, the number of stacked inter- and intra-modality
transformers is $N_{o \leftrightarrow s}=5$, $N_{o \leftarrow
o}=5$ and $N_{s \leftarrow s}=9$. The encoder hidden's vectors are of dimension
$d=768$ and each attention layer has $H{=}12$ heads.
Moreover, to reduce computation, we set the maximum sentence length to $N_s=20$
tokens and the number of visual objects to $N_o=36$.
Training is done on four V100 GPUs. 

For NLVR2 \cite{suhr2019corpus}, we finetune during 4 epochs using Adam
optimizer \cite{kingma2014adam}. The learning rate is set to $5*10^{-5}$ and the batch
size is 32.
We only supervise with the task-specific binary
objective, \ie we drop all the supervision signals used for encoder training. 
% add more evaluation with other metrics.
%DO not compare LXMERT with ours finetune. But emphasize the fact that we achieve
%a similar accuracy without finetuning on both datasets.

% TABLE: raw vqa gqa
\begin{table}
   \begin{center}
   {\caption{Impact of the proposed object-word alignment weak supervision on the VQA task.
   The presented results are calculated on the GQA~\cite{Hudson_2019_CVPR} test-std split.}\label{table:VQA_RAW}}
   \begin{tabular}{lcc}
   \hline
   \rule{0pt}{12pt}
   Models&Consistency&Accuracy\\
   \hline
   \\[-6pt]
   ours (w/o alignment supervision) & 79.5 & 54.9\\
   \textbf{ours (with alignment supervision)} & \textbf{89.7} & \textbf{60.5}
   \\
   \hline
   \\[-6pt]
   \end{tabular}
   \end{center}
\end{table}

\myparagraph{Visual Question Answering}
Table~\ref{table:VQA_SOTA} compares the results of applying our vision-language encoder on the VQA task versus the recent published works.
As one may observe, our model obtains the 2nd-best SOTA-result, just after the
NSM model~\cite{hudson2019learning}. The latter is fundamentally different from our
approach (contrary to NSM, our model operates on raw images rather than on the
scene graphs).
Moreover, it is important to highlight that, unlike previous work, our model has
not been finetuned on the target dataset making the obtained result even more
significant.

% Like our method, LXMERT vision-language encoder is also supervised with visual question answering signal.
% Thus we evaluate LXMERT without finetuning it on the downstream task following the same method as with our model.
% This direct evaluation teaches us on the quality of the vision-language encoder
% representation.
In order to quantify the impact of the our object-word alignment weak supervision on the VQA task, we evaluate the two versions of our model, with and without the proposed loss, on the GQA dataset.
The results are reported in Table~\ref{table:VQA_RAW}.
One may observe that the proposed weak supervision boosts the accuracy with $+5.6$ points.
Moreover, when we focus on the metric which explicitly measures the reasoning capacity of the model, namely the consistency, our weakly-supervised alignment allows to gain more than $+10$ points.
This demonstrates that, by enforcing the model to explicitly align words with visual objects, we obtained a finer multimodal representation.

\myparagraph{Natural Language for Visual Reasoning (NLVR2)}
% Comparison with SOTA. Add UNITER
%VisualBERT & 67.4\% & 67.0\% & - \\
\begin{table}
   \begin{center}
   {\caption{Evaluation of the proposed object-word alignment weak supervision on the NLVR2 evaluation splits. Models marked with * have been
   ran by the authors of \cite{suhr2019corpus}.}\label{table:NLVR_SOTA}}
   \begin{tabular}{lcc}
   \hline
   \rule{0pt}{12pt}
   Models&Dev.&Test-P\\
   \hline
   \\[-6pt]
   MAC* \cite{hudson2018compositional} & 50.8 & 51.4 \\
   FiLM* \cite{perez2018film} & 51.0 & 52.1 \\
   CNN+RNN* \cite{suhr2019corpus} & 53.4 & 52.4 \\
   MaxEnt \cite{suhr2019corpus} & 54.1 & 54.8 \\
   LXMERT \cite{tan2019lxmert} & 74.9 & 74.5 \\
   \hline
   \rule{0pt}{8pt}
   \textbf{ours} & \textbf{75.8} & \textbf{75.5}
   \\
   \hline
   \\[-6pt]
   \end{tabular}
   \end{center}
\end{table}
As shown in Table~\ref{table:NLVR_SOTA}, our method outperforms the
published\footnote{At the time of the writing, an unpublished work, called
UNITER~\cite{chen2019uniter}, reported a better result on NLVR2.} SOTA accuracy on
NLVR2 with a gain of $+1$ point.
Furthermore, we have performed the same ablation analysis as for the VQA task
(i.e. with and without the object-word alignment weak supervision), and the
obtained results are summarized in Table~\ref{table:NLVR_CMPR}.
These results are coherent with those calculated on the VQA task confirming the advantage of the proposed supervision.
Note that the scores in Table~\ref{table:NLVR_CMPR} are reported both for unbalanced and balanced subsets of the NLVR2 dataset.
This split takes into account the visual biases present in the dataset.
The benefit of our fine-grained alignment supervision method is constant between both subsets, showing that the gain is not bias-related.

%  Comparison with LXMERT (using balanced/unbalanced)
\begin{table}
   \begin{center}
   {\caption{Impact of the proposed object-word alignment weak supervision on the Visual Reasoning grounded by Language task.
   The presented results are calculated on the Test-P part of the NLVR2 dataset.}\label{table:NLVR_CMPR}}
   \begin{tabular}{lccc}
   \hline
   \rule{0pt}{12pt}
   Models&Test-P&Unbalanced&Balanced\\
   \hline
   \\[-6pt]
   ours (w/o alignment supervision) & 74.5\% & 76.0\% & 73.1\%\\
   \textbf{ours (with alignment supervision)} & \textbf{75.5\%} & \textbf{77.2\%} & \textbf{74.5\%}
   \\
   \hline
   \\[-6pt]
   \end{tabular}
   \end{center}
\end{table}

\section{Visualizing Reasoning}
\label{sec:visu}
\begin{figure*}[ht!]
   \begin{center}
      \includegraphics[width=1.0\linewidth]{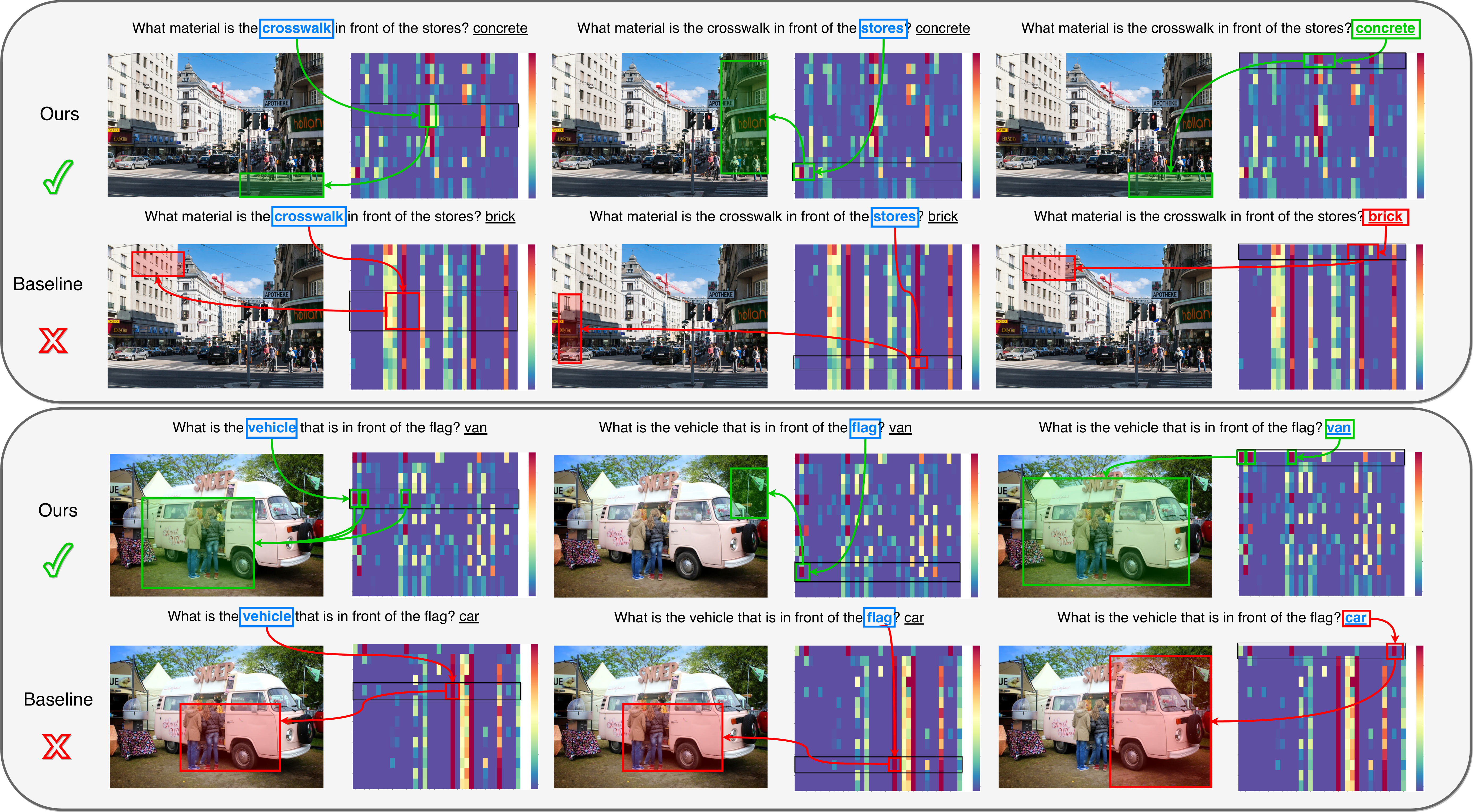}
   \end{center}
      \caption{
      Visualization of the attention maps of the penultimate (=4th) inter-modality transformer.
      Word-object alignment does not emerge naturally for the baseline
      (without object-word alignment supervision), whereas our model with the
      proposed weakly-supervised objective learns to pay strong
      cross-attention on
      co-occurring combinations of words and objects in the
      scene. In the attention maps, rows represent words and columns represent
      visual objects. For the sake of visibility, we display the bounding box of
      the detected object with the highest activation regarding to the selected
      word. The predicted answer (underlined) is written after the question. Its
      corresponding language token is $[CLS]$,  \ie the first row in attention maps.}
   \label{fig:attention_visu}
\end{figure*}

The reasoning capabilities of high-capacity deep neural networks are notoriously
difficult to interpret, as inputs and intermediate activations are embedded in
high-dimensional spaces in non trivial applications. Vision-language models are
no exceptions, therefore we propose visualizations of some of the key
activations of the proposed model.
%As to illustrate the benefit of the object-word alignment, we propose to visualize what has been learned by the encoder.
%These qualitative experiments provide some insights about the reasoning mechanisms induced by the training supervision.
Such visualizations --- when wisely chosen --- can be a step toward more interpretable models,
especially in the field of visual reasoning, where distinguishing real reasoning
(i.e. which follows causal chains) from educated guesses (i.e. exploiting
subtle statistic biases in the data) can be difficult. The visualizations in
this Section are obtained from the dev-test split of the GQA \cite{Hudson_2019_CVPR} dataset.

% \subsection{Visualizing cross-modality attention maps}
We inspect the attention maps inside the inter-modality transformers, which
illustrates the information flow between the two modalities (vision and language).
Generally, attention maps convey information on the importance that a neural map
poses on local areas in input or activations. In the particular case of our
model, the inter-modalilty attention map visualize how modalities are fused by
the model, as they give weight to outputs for a given word as a function of a given
object (or vice-versa).
% These maps indicate Thereby, it informs us on the exploration mechanism
% set by the model to achieve the reasoning.

Following equation (\ref{eq:attention_maps}), 
we visualize the values $\alpha_{ij}$, showing the attention given to the pair $(s_i, o_j)$.
We visualize the map of the 4$^{th}$ inter-modality transformer and sum the maps
over the $12$ parallel attention heads, comparing the maps of our proposed model
with and without the proposed object-word alignment supervision in Figure~\ref{fig:attention_visu}. 

The effectiveness of the new object-word alignment objective is corroborated 
by attention units which are higher for object-word pairs referring to the same entity in our model.
% Thus, we can see that during the training the vision-language encoder has learned to align both modalities
% in order to let the information flow only between tokens pointing to the same high level entity. 
We observe a radically different behavior in the baseline's attention maps,
where attention is less-fine grained: 
roughly uniform attention distributions indicate that the layer outputs of all
words attend to roughly the same objects.

\textbf{Caveat:} we do not want to imply, that the exact word-object alignment
in the inter-modality layer is indispensable for a given model to solve a
reasoning task, as a complex neural network can model relationships in
the data in various different layers. However, we do argue, that some form of
word-object alignment is essential for solving vision-language tasks, as the
model is required to query whether concepts from the question are present in the
image, and eventually query their relationships to other concepts. Inductive
bias has been added to model for this type of reasoning in the form of
inter-modality layers, and it is therefore natural to inspect whether this
cross-attention emerges at this exact place. We would also like to point out
that we do not force or favor word-object alignment at a specific layer, as our
proposed objective is injected through a new module attached to the
inter-modality layer (see Figure~\ref{fig:overall}). The attention maps show
that the objective is successfully propagated from the new alignment head to the
inter-modality layer.
\section*{Conclusion}
In this work, we design a vision-language encoder and train it with a novel
object-word alignment weak supervision.
To this end, we carefully design a soft alignment signal taking into account
both spatial and semantic alignment between the words and the detected visual objects.
We empirically show the benefit of this new
supervision on two tasks requiring to reason over images, namely the VQA and the
Language-driven Comparison of Images on which
we obtain the SOTA-level accuracies.
We also provide a qualitative visualization of the attention distributions of our
model, showing that attention units are higher for object-word
pairs referring to the same entity, and that the proposed object-word alignment
does not emerge naturally without supervision.
Future work will explore the application of this weak supervision signal to
other vision-language tasks, including image retrieval.

% \begin{figure}[t]
% \begin{center}
%    %\includegraphics[width=0.8\linewidth]{egfigure.eps}
% \end{center}
%    \caption{Example of caption.  It is set in Roman so that mathematics
%    (always set in Roman: $B \sin A = A \sin B$) may be included without an
%    ugly clash.}
% \label{fig:long}
% \label{fig:onecol}
% \end{figure}

% \ack We would like to thank the referees for their comments, which
% helped improve this paper considerably

\bibliography{ecai}
\end{document}